%% file: main.tex
\documentclass[10pt,twocolumn,letterpaper]{article}

\usepackage{cvpr}              

\usepackage{makecell}
\usepackage{multirow}


\definecolor{cvprblue}{rgb}{0.21,0.49,0.74}
\usepackage[pagebackref,breaklinks,colorlinks,allcolors=cvprblue]{hyperref}
\usepackage[accsupp]{axessibility}  

\def\paperID{xxxxx} 
\def\confName{CVPR}
\def\confYear{2026}

\newcommand\nonumfootnote[1]{%
\begingroup%
    \renewcommand\thefootnote{}\footnote{\hspace{-3.7pt}#1}%
    \addtocounter{footnote}{-1}%
\endgroup%
}

\title{StreamDiT: Real-Time Streaming Text-to-Video Generation}

\author{Akio Kodaira$^{1,2*}$ 
\hspace{0.04cm}
Tingbo Hou$^{2,*}$ \hspace{0.02cm}
Ji Hou$^2$ \hspace{0.02cm}
Markos Georgopoulos$^2$ \hspace{0.02cm} \\
Felix Juefei-Xu$^2$ \hspace{0.02cm}
Masayoshi Tomizuka$^1$
Yue Zhao$^{2,*}$\\
\\
$^1$UC Berkeley \hspace{0.02cm}
$^2$Meta \vspace{0.1cm}\\
}

\begin{document}
\twocolumn[{
\maketitle
\begin{center}
    \vspace{-6mm}
	\centering
    \small
	\captionsetup{type=figure}
 \includegraphics[width=\textwidth]{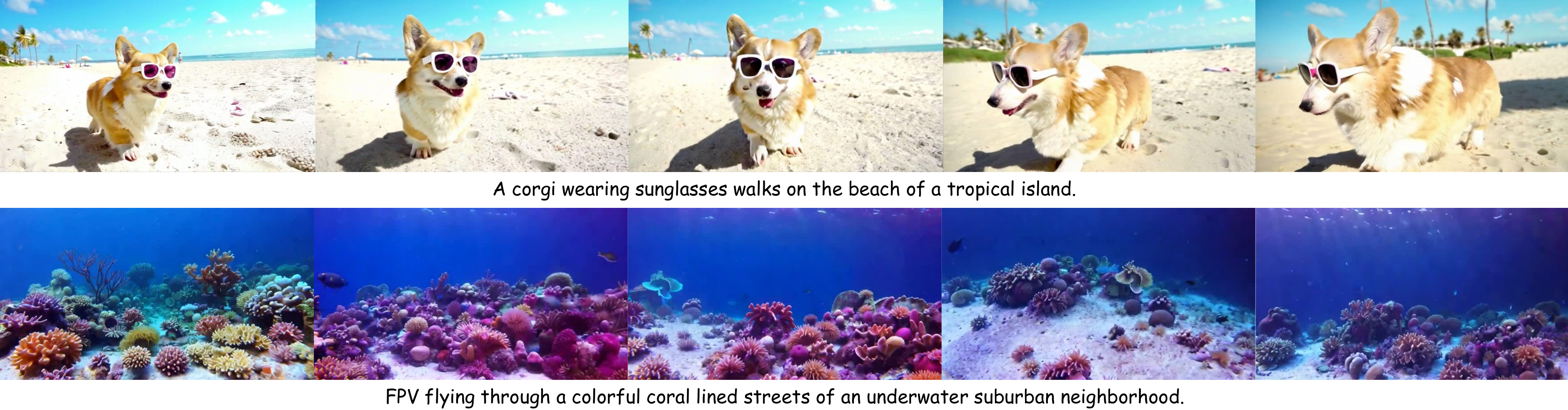}
    \captionof{figure}{StreamDiT-4B: video generation can be streaming and real-time.}
    \vspace{1mm}
    \label{fig:teaser}
\end{center}
}]
\maketitle
\input{sec/0_abstract}
\input{sec/1_intro}
\input{sec/2_related}
\input{sec/3_buflow}
\input{sec/4_model}
\input{sec/6_exp}
\input{sec/7_conclu}
\input{sec/8_acknowledge}

{
    \small
    \bibliographystyle{ieeenat_fullname}
    \bibliography{main}
}

\input{sec/Appendix}

\end{document}

%% file: sec/0_abstract.tex
\begin{abstract}
\nonumfootnote{\noindent $^*$ denotes equal contributions}
\nonumfootnote{This work was done while the first author was an intern at Meta.}
Recently, great progress has been achieved in text-to-video (T2V) generation by scaling transformer-based diffusion models to billions of parameters. However, existing models typically produce only short clips offline, restricting their use cases in interactive and real-time applications. This paper addresses this challenge by proposing StreamDiT, an efficient trainable solution for streaming video generation. StreamDiT training modifies flow matching by adding a moving buffer with progressive denoising. We design mixed partitioning schemes of buffered frames to boost both content consistency and visual quality. StreamDiT modeling is based on adaLN DiT with varying time embedding and window attention. In addition, we propose a multistep distillation method tailored for StreamDiT. Sampling distillation is performed in each segment of a chosen partitioning scheme. After distillation, the total number of function evaluations (NFEs) is reduced to the number of chunks in a buffer. We apply StreamDiT to a 4B-parameter model, which reaches real-time performance for 512p resolution at 16 FPS on one GPU. We evaluate our method through both quantitative metrics and human evaluation. Our model enables real-time applications, \eg streaming generation, interactive generation, and video-to-video editing. More video results are available at the project website: \href{https://cumulo-autumn.github.io/StreamDiT/}{https://cumulo-autumn.github.io/StreamDiT/}.
\vspace{-15pt}
\end{abstract}

%% file: sec/1_intro.tex
\section{Introduction}
\label{sec:intro}

Video generation is a trending problem in computer vision, with rapid progress in recent years. Diffusion models have been adopted for text-to-video (T2V) generation, and demonstrated capabilities of generating high-quality videos, with a potential as world simulators. The success was built upon the scalability of transformer-based architectures and huge compute resources to train billions of parameters. Despite achievements, many existing models are designed to generate short video clips, due to the increasing cost after scaling. As a result, generating long videos with low latency remains an extreme challenge~\cite{li2024surveylongvideogeneration}, which is required in interactive and real-time applications.

Modern T2V models~\cite{moviegen,hunyuan,step-t2v} are based on Diffusion Transformers (DiTs)~\cite{Peebles2022DiT} with bidirectional attention.
Increasing video length is expensive due to the quadratic complexity of transformers with respect to sequence length. To make diffusion models capable of generating long videos, some approaches~\cite{xie2024progressiveautoregressivevideodiffusion,yin2025slowbidirectionalfastautoregressive,huang2025selfforcing} attempted to combine autoregressive (AR) and diffusion models, by adding progressive noise or causal masking. In an AR setup, a generation frame can only see previous context frames in attention, posing a limitation on its quality. As shown in~\cite{li2024autoregressiveimagegenerationvector}, replacing causal attention with bidirectional attention leads to a massive quality gain. Modern video generation models~\cite{opensora,moviegen,hunyuan} are also based on bidirectional attention. On another axis, AR frame-by-frame prediction has low streaming latency. However, it still needs to wait for a full denoising process to obtain a clean output frame. Diffusion with bidirectional attention can have high throughput, but its latency is much lower than AR. These observations lead us to seek a solution with low latency, high throughput, and high quality for streaming video generation. 

We are inspired by non-trainable streaming generation \eg StreamDiffusion~\cite{kodaira2023streamdiffusionpipelinelevelsolutionrealtime} and FIFO-Diffusion~\cite{Kim2024FIFO}. StreamDiffusion utilizes pre-trained image diffusion models to edit an input video stream via batch denoising. It suffers from a lack of frame-to-frame consistency, leading to visual artifacts when generating continuous video streams. FIFO-Diffusion enqueues a sequence of frames with different noise levels. It pops up a clear frame after one denoising step and enqueues a noise frame in the queue. However, they do not come with any training solution, which limits their quality. Another missing effort is sampling distillation, which is crucial for real-time applications. Since the queue is updated at every denoising step, popular distillation methods \eg step distillation~\cite{guideddistillation} and consistency distillation~\cite{lcm} cannot be applied directly.

To address the challenges, we propose a complete solution for streaming video generation including training, modeling, and distillation, named StreamDiT. For training, we modify the flow matching~\cite{lipman2023flowmatching} by introducing a moving buffer of frames. Within a buffer, we adopt full attention on latent frames so that our method can be seamlessly applied to modern T2V models. At inference, the buffer is moving along the frame dimension, generating clean frames sequentially. Similar to previous work~\cite{xie2024progressiveautoregressivevideodiffusion,Kim2024FIFO}, we allow buffered frames with different noise levels. We design a unified partitioning of latent frames with different chunk sizes and micro steps. The uniform noise~\cite{lipman2023flowmatching} and the diagonal noise~\cite{Kim2024FIFO,xie2024progressiveautoregressivevideodiffusion} are special cases of our partitioning. With mixed training of different schemes, we can enhance video consistency and avoid overfitting on a specific scheme.

For modeling, we design our StreamDiT model architecture to fit its training process while being efficient for real-time applications. We modify adaLN DiT~\cite{Peebles2022DiT} by adding varying time embedding and replacing full attention with window attention~\cite{swintransformer}. We first train our StreamDiT as a basic T2V model, and then adapt it to streaming video generation. To achieve real-time applications, we also design a multistep distillation tailored for StreamDiT with a chosen partitioning scheme. With that, we distill our model to 8 sampling steps without classifier-free guidance (CFG). Finally, our model can generate video frames at 16 FPS on one H100 GPU, achieving real-time performance. ~\cref{fig:teaser} shows some video results of our real-time streaming generation.

Our contributions are summarized as follows:
\begin{itemize}
    \item We propose StreamDiT training for streaming video generation, which is based on flow matching by introducing a moving buffer. We articulate a generalized partitioning of buffered frames that covers uniform noise, diagonal noise, and others in between. We also point out a requirement for generation models such that time embedding needs to be separable in the frame dimension. 
    \item We design StreamDiT model as adaLN DiT with varying time embedding and window attention. Our StreamDiT has 4B parameters and is trained to generate videos at 512p resolution. We show that using mixed training with different partitioning schemes can improve the quality and flexibility of streaming video generation.
    \item We build a real-time solution using multistep distillation tailored for StreamDiT. Choosing a partitioning scheme that balances inference speed and quality, we distill micro steps to one for each segment. With all the efforts combined, our distilled model achieves real-time performance with 16 FPS on one GPU. In addition, we show some results to highlight potential real-time applications using our model.
\end{itemize}

%% file: sec/2_related.tex
\section{Related Work}

\subsection{Text-to-Video Generation}

Arising with the development of text-to-image generation (T2I)~\cite{dai2023emu}, video generation bursts by introducing temporal modules.
Earlier work~\cite{zhou2023magicvideo,he2023latentvideodiffusionmodels,blattmann2023alignlatents,guo2023animatediff,chen2024videocrafter2} was based on UNet structures with temporal layers added to T2I models. As a pivot, DiT~\cite{Peebles2022DiT} demonstrated that transformer architecture is more scalable than UNet architecture for diffusion models. VDT~\cite{Lu2024VDT} proposes to use separated spatial and temporal attentions in a transformer block. GenTron~\cite{chen2023gentron} introduces a family of transformer-based diffusion models for image and video generation. Snap Video~\cite{menapace2024snapvideo} proposes a transformer architecture with joint spatiotemporal blocks, which is trained faster than U-Nets. ViD-GPT~\cite{gao2024vidgpt} uses causal attention in temporal domain and frames as prompts, following the design of GPT in LLMs. Due to the limitations of compute, these models are trained at small or medium scales, up to a few of billion parameters, generating videos of a few seconds. In the latest work of MovieGen~\cite{moviegen}, the DiT model has been scaled to 30B parameters and can generate videos up to 16 seconds, with high motion and aesthetic qualities.

In parallel, the open-source community is also growing fast. OpenSora~\cite{opensora} released a small T2V model with about 1B parameters, using factorized transformers for spatial and temporal domains. Hunyuan~\cite{hunyuan} video model reaches 13B parameters, which uses MMDiT~\cite{esser2024sd3} with concatenated text and video tokens in self-attention. For auto-encoder, it adopts causal 3D convolution~\cite{yu2024languagemodelbeatsdiffusion} in its VAE. Step-Video-T2V~\cite{step-t2v} with 30B parameters is a large open-source model for video generation, which reduces the model size gap between open-source and closed-source.

\subsection{Extensible Video Generation}
The literature of extensible long video generation can be roughly categorized as training and training-free methods. Training new models or new modules injected to existing models requires a large amount of compute resources. NUWA-XL~\cite{yin2023nuwaxl} proposes a diffusion over diffusion architecture to generate long videos in a coarse-to-fine manner.
StreamingT2V~\cite{henschel2024streamingt2v} proposes an autoregressive approach for long video generation, using a short-term memory block and long-term memory block of previous chunks. In~\cite{xie2024progressiveautoregressivevideodiffusion}, a similar method was proposed by assigning latent frames with progressively increasing noise levels. VideoTetris~\cite{tian2024videotetris} trains a ControlNet branch of condition frames for long video generation. A recent work by Loong~\cite{wang2024loong} proposes an autoregressive LLM-based approach for video generation. The model is trained on 10-second videos and can be extended to one-minute long. To reduce the training and inference gap of AR video generation, self-forcing~\cite{huang2025selfforcing} was proposed, which uses student predictions as previous frames in the context. To reduce the streaming latency, it generates one latent frame at a time. The current AR models are hard to compete with full-attention diffusion models on quality.

For training-free methods, Gen-L-Video~\cite{wang2023genlv} discovers that the denoising path of a long video can be approximated by joint denoising of overlapping short videos in the temporal domain. MimicMotion~\cite{zhang2024mimicmotion} extends MultiDiffusion~\cite{bar2023multidiffusion} to the temporal domain. It blends frames within overlap region progressively along the denoising process. Video-Infinity~\cite{tan2024videoinf} distributes a long-form video task to multiple GPUs, with dual-scope attention that modulates temporal self-attention to balance local and global contexts efficiently across the devices. FreeNoise~\cite{qiu2024freenoise} proposes a simple solution to generate consistent long videos by rescheduling a sequence of noises for long-range correlation and performing temporal attention over them by window-based function. Reuse-and-Diffuse~\cite{gu2023reusediffuse} proposes an iterative denoising for T2V, with staged guidance. It reuses intermediate denoising results from the previous clip to condition the current clip. FIFO-Diffusion~\cite{Kim2024FIFO} proposes an inference technique that iteratively performs diagonal denoising, which concurrently processes a series of consecutive frames with increasing noise levels in a queue.

\subsection{Sampling Efficiency}
Step distillation has been widely adopted for accelerating diffusion models. Compared to advanced numerical solvers, sampling distillation demonstrates better quality with fewer number of sampling steps. Progressive distillation~\cite{salimans2021progressive} distills two denoising steps of a teacher model to one step of a student model. Guided distillation~\cite{guideddistillation} proposed to first distill two function evaluations of CFG to one, and then perform progressive distillation on sampling steps. Consistency distillation~\cite{song2023consistencymodels,lcm} maps any points on a diffusion trajectory to the same origin, and therefore reduces the number of steps needed for sampling. Multistep consistency models~\cite{heek2024multistepconsistencymodels} splits the diffusion process to multiple segments, and consistency distillation can be applied at each segment. By allowing more budget on sampling steps, Multistep consistency distillation can achieve higher quality.

To further improve sampling efficiency, one-step approaches have been proposed, including UFOGen~\cite{ufogen} and DMD~\cite{dmd}. The idea is to directly match distributions without the Gaussian assumption. They require to host auxiliary models (discriminator or fake score networks) during training, posing challenges for applying to large video models.

%% file: sec/3_buflow.tex
\section{StreamDiT Training}

\subsection{Buffered Flow Matching}
\label{sec:theory}

\textbf{Flow Matching:} Our training is based on Flow Matching (FM)~\cite{lipman2023flowmatching}.
FM produces a sample from the target data distribution by progressively transforming a sample from an initial prior distribution, such as a Gaussian.
For a data sample $\mathbf{X}_{1}$, FM samples a time step $t \in [0, 1]$, and noise $\mathbf{X}_{0} \sim \mathcal{N}(0, 1)$. These are used to create a training sample $\mathbf{X}_t$.
FM predicts the velocity $\mathbf{V}_t$ that moves the sample $\mathbf{X}_t$ in the direction of data sample $\mathbf{X}_1$.

A simple linear interpolation or the optimal transport (OT) path~\cite{lipman2023flowmatching} is used to construct $\mathbf{x}_t$, \ie
\begin{equation}
  \mathbf{X}_t = t~\mathbf{X}_1 + (1 - (1-\sigma_{\mathrm{min}})t)~\mathbf{X}_0,
\end{equation}
where $\sigma_{\mathrm{min}}$ is the standard deviation of $x$ at $t=1$.
Thus, the ground truth velocity can be derived as
\begin{equation}
    \mathbf{V}_t = \dfrac{d\mathbf{X}_t}{dt} = \mathbf{X}_{1} - (1-\sigma_{\mathrm{min}}) \mathbf{X}_{0}.
\end{equation}
It is worth noting that this target is irrelevant to timestep $t$.
With parameters $\Theta$ and text prompt $\mathbf{P}$, the predicted velocity is written as $u(\mathbf{X}_t, \mathbf{P}, t; \Theta)$, and the training objective is represented as
\begin{equation}
  \label{eq:fm}
  \mathbb{E}_{t, \mathbf{X}_t}\|u(\mathbf{X}_t, \mathbf{P}, t; \Theta) - \mathbf{V}_t\|^2.
\end{equation}

At inference, an FM model predicts the velocity to a clean sample on every denoising step. With the Euler solver, the inference can be formulated as 
\begin{equation}
\label{eq:denoise}
    \mathbf{X}_{t+\Delta t}=\mathbf{X}_t+u(\mathbf{X}_t, \mathbf{P}, t; \Theta) \Delta t,
\end{equation}
where $\Delta t$ is the step size.

\textbf{Buffered Flow Matching:} We consider streaming video generation as a sequence of (possibly latent) frames $[f_1, f_2, \dots, f_N]$, and $N$ can be infinite. For a video diffusion model with frame buffer $B$, the clean data sample starting with frame $f_i$ is denoted as $\mathbf{X}_1^i=[f_{i+1},\dots,f_{i+B}]$. We allow different noise levels to the frames: $\tau=[\tau_1,\dots,\tau_B]$ as a monotonically increasing sequence. Thus a training example can be constructed as
\begin{equation}
  \mathbf{X}_{\tau}^i = \tau \circ ~\mathbf{X}_1^i + (1 - (1-\sigma_{\mathrm{min}}) \tau) \circ ~\mathbf{X}_0,
\end{equation}
where $\circ$ denotes element-wise product. Please note that the noise sample $X_0$ remains the same.

At inference, the buffer is updated by model predicted flow
\begin{equation}
\label{eq:bfdenoise}
    \mathbf{X}_{\tau+\Delta\tau}^i=\mathbf{X}_{\tau}^i+u(\mathbf{X}_{\tau}^i, \mathbf{P}, \tau; \Theta) \circ \Delta \tau,
\end{equation}
where $\Delta \tau$ is a sequence of step sizes. If one or more frames achieve the final denoising step, they are popped out of the buffer, and new noise frames are pushed at the beginning of the buffer. Therefore, it can generate streaming video sequences.

\subsection{Partitioning Scheme}
\label{sec:buflow}

\begin{figure}
    \centering
    \includegraphics[width=\linewidth]{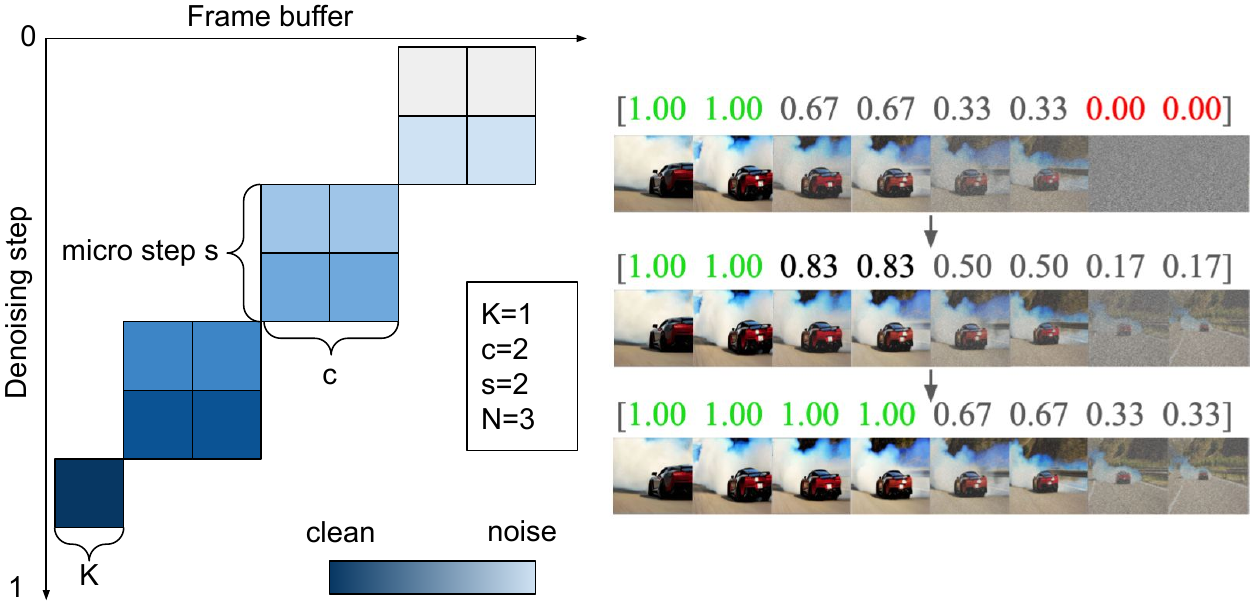}
    \caption{Illustration of StreamDiT partitioning. We partition the buffer to $K$ reference frames and $N$ chunks. Each chunk has $c$ frames and $s$ micro denoising steps.}
    \label{fig:buflow}
\end{figure}

To facilitate flexible training and inference of StreamDiT, we design a unified partitioning of buffered frames. As illustrated in~\cref{fig:buflow}, the buffer is partitioned to $K$ reference (context) frames and $N$ chunks. Each chunk has $c$ frames and $s$ micro denoising steps.

\textbf{Context Reference Frames:}
To enhance temporal consistency, we can optionally cache the last $K$ fully denoised frames at the beginning of the buffer. We refer to these clean frames as reference frames. They participate in the denoising step as input but are no longer denoised. The reference frames are updated in the same way as other frames when the buffer moves. Allowing optional reference frames matches the design of FIFO-Diffusion~\cite{Kim2024FIFO}, which can be viewed as a special case. Since our method is trainable, we found that reference frames can be skipped for streaming video generation. Hence, we set $K=0$ in the rest of our work.

\textbf{Chunked Denoising:}
Instead of denoising frame by frame, we group frames into stream chunks, with each chunk containing a specified number of frames indicated by chunk size. Noise levels are now applied at the chunk level, and each time a chunk of frames exits the pipeline.
Let $N$ denote the number of stream chunks and $c$ the number of frames in each chunk. Then the total number of frames processed at any time is $K + N \times c$, and the number of denoising steps is constrained to $N$. 

\textbf{Micro Step:}
For better performance, additional denoising steps are usually helpful. A straightforward approach to address this limitation is to increase the size of the buffer, which will increase the serving cost, and therefore needs to be bounded by a computation budget.

To overcome this, we introduce an additional dimension of the design called micro-denoising step, illustrated in~\cref{fig:buflow}. The core idea of micro step is to denoise stream chunks along the temporal axis while stagnating at a fixed spatial position for a certain amount of time. Let $s$ denote the micro step; then each stream chunk undergoes $s$ denoising steps before advancing to the next noise level and moving toward the output. This modification effectively extends the total number of denoising steps to  $s \times N$  without increasing the buffer size.

With the incorporation of reference frames, chunked frames, and micro-step denoising, the following equations hold:
\begin{equation}
\begin{split}
B &= K + N \times c \\
T &= s \times N
\end{split}
\label{eq:pipeline}
\end{equation}
where $B$ is the total length of the frame fed into the model,  $N$ is the number of stream chunks,  $c$ is the number of frames in each stream chunk, and $T$ represents the total number of inference denoising steps.

\begin{table}[]
    \centering
    \resizebox{\columnwidth}{!}{
    \begin{tabular}{c|c|c|c}
    \toprule
         Method & Scheme & Consistency & Streaming \\
         \midrule
         Uniform & $c=B,s=1$ &High & No \\
         \midrule
         Diagonal & $c=1,s=1$& Low & Yes \\
         \midrule
         StreamDiT & $c=[1,\dots,B], s=\frac{T}{N}$ & High & Yes\\
         \bottomrule
    \end{tabular}
    }
    \caption{StreamDiT unifies different partitioning schemes.}
    \label{tab:buflow}
\end{table}

\textbf{Mixed Training:} 
As shown in~\cref{tab:buflow}, StreamDiT unifies partion schemes from bidirectional attention with uniform noise to progressive diagonal noise~\cite{Kim2024FIFO,xie2024progressiveautoregressivevideodiffusion}. The latter enables streaming generation, but hurts consistency of generated content. To enhance consistency and avoid overfitting, we adopt mixed training with different schemes. This drives the model to learn generalized denoising with different noise levels, instead of memorizing fixed noise levels. It is worth noting that our mixed training covers diffusion and FM training. Therefore, our model can work as a standard T2V generation without streaming. This is also used to initialize our streaming generation.

According to~\cref{fig:buflow}, frames in each chunk correspond to a distinct segment of the overall time step range. Thus, during training, we sample a random time step for the $i$-th chunk as follows:
\begin{equation}
\tau_i \sim \text{Uniform}\left(\left[\frac{T}{N} \cdot (i-1), \frac{T}{N} \cdot i\right]\right)
\label{eq:stream_t}
\end{equation}
Interestingly, the StreamDiT training can be viewed as parallel training of the full range of denoising. 

%% file: sec/4_model.tex
\section{StreamDiT Modeling}

\subsection{Model Architecture}
\label{sec:model}
\begin{figure}
    \centering
    \includegraphics[width=0.8\linewidth]{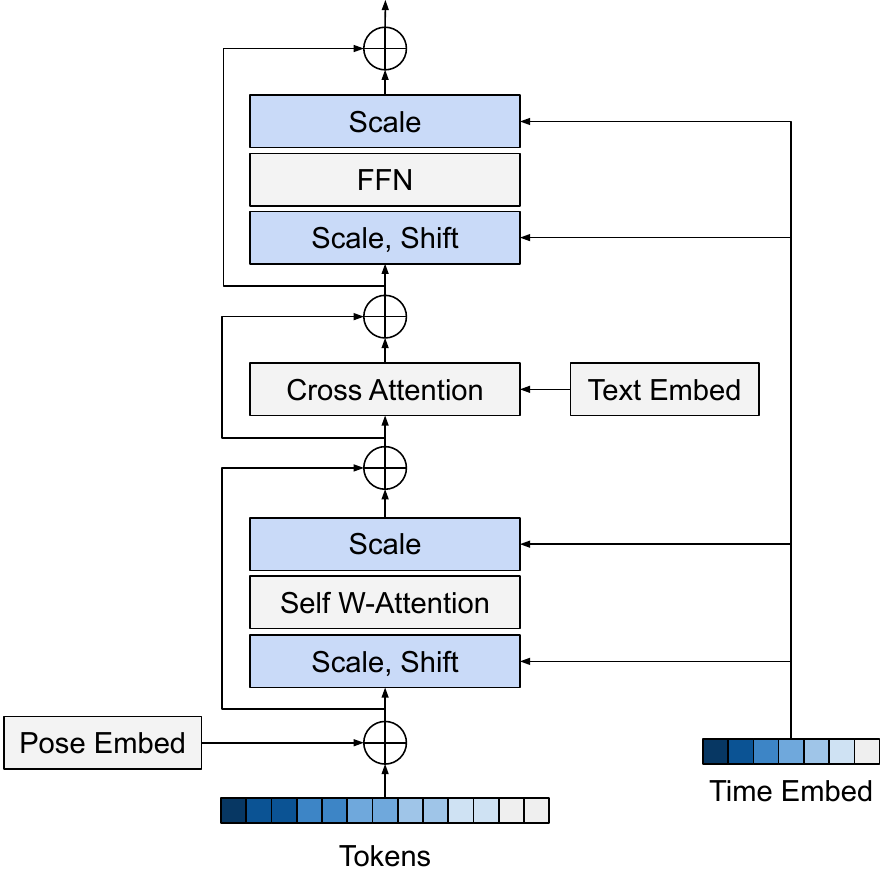}
    \caption{StreamDiT with varying time embedding and window attention. We modified the standard adaLN DiT with varying time embeddings applied to scale and shift modulations.}
    \label{fig:tdit}
\end{figure}

\textbf{Time-Varying DiT:}
As shown in~\cref{sec:theory}, our StreamDiT changes the scalar condition $t$ to a sequence $\tau$ in the model. This requires that the time condition should be separable in the frame dimension. We follow the standard adaLN DiT~\cite{Peebles2022DiT} architecture, where time embedding is used for scale and shift modulations, and modify it with varying time embedding. Specifically, we reshape the latent tensor to 3D: $[F,H,W]$, and apply time embedding along the first dimension. As shown in~\cref{fig:tdit}, we modify the adaLN DiT with varying time embedding. Input tokens are also corrupted with different levels of noise. This architecture matches the design of our StreamDiT.

\begin{figure}
    \centering
    \includegraphics[width=0.8\linewidth]{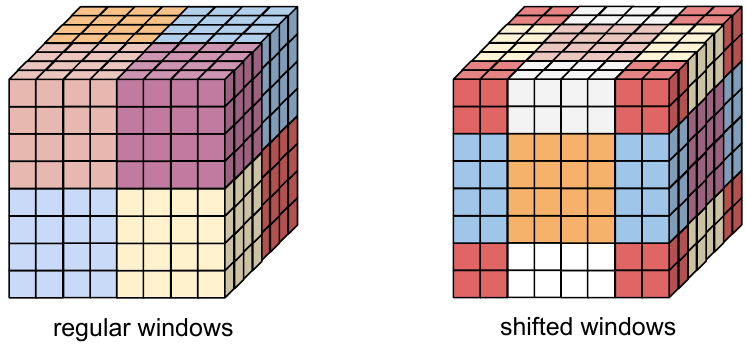}
    \caption{Illustration of window partitioning: regular windows and shifted windows.}
    \label{fig:window}
\end{figure}

\textbf{Window Attention:}
To make the DiT model more efficient, we adopt window attention~\cite{swintransformer} for all self-attention layers. Specifically, we partition a 3D latent with shape $[F, H, W]$ to non-overlapped windows with size $[F_w, H_w, W_w]$, and apply masking when computing self-attention, such that a token can only see tokens within the same window. To ensure consistency across windows, we shift by half of the window size every other layer. As illustrated in~\cref{fig:window}, voxels at axis ends will be warped over to the beginnings in the shifting. Accumulatively, global token communication can broadcast to all tokens, while each computation is efficient with local attention. The complexity of window attention is only $\frac{F_w\times H_w\times W_w}{F\times H\times W}$ of full attention.

\textbf{Other Components:}
Since we are targeting real-time applications, we chose a moderate size (4B) for our DiT. We reuse a temporal auto-encoder (TAE) in Movie Gen~\cite{moviegen} with compression rate 4 in temporal domain and 8 in spatial domain. The latent channel size is 8, with the consideration that a moderate latent space is easier to learn for small generation models. The trade-off is that with a smaller temporal compression rate, the model needs to generate more latent frames to reach the same FPS. 

For text encoders, we adopt the same ones from Movie Gen~\cite{moviegen}, including UL2~\cite{tay2023ul2unifyinglanguagelearning}, ByT5~\cite{xue2022byt5tokenfreefuturepretrained}, and MetaCLIP~\cite{xu2024demystifyingclipdata}. The text encoders run fast on GPU. We only run text encoders when the prompt is changed for streaming video generation. Therefore, its computation time is negligible. 

\subsection{Multistep Distillation}
\label{sec:distill}
Sampling distillation is a key component to build real-time streaming video generation. Due to the modifications of frame partitioning and micro denoising steps made by StreamDiT, standard sampling distillation methods~\cite{salimans2021progressive,guideddistillation,heek2024multistepconsistencymodels} cannot be applied. Thus, a customized sampling distillation for StreamDiT is needed. 

As described in~\cref{sec:buflow}, StreamDiT partitions frames in a buffer to $N$ chunks, each of which has $s$ denoising steps before moving the buffer. In particular, we aim at reducing the micro-step $s$ in~\cref{eq:pipeline}. Our StreamDiT model is trained with mixed partitioning schemes, to make it flexible to different scenarios. In distillation, we first choose a partitioning scheme. Specifically, we set $K=0$, $c=2$, $s=16$, $N=8$ for the teacher model. This indicates that the teacher model has $s*N=128$ denoising steps with CFG. The FM trajectory is split to $N$ segments. We then perform step distillation in each segment separately. In practice, we perform both step and guidance distillation at the same time, by distilling multiple CFG steps of the teacher into a single conditional forward pass of the student.

For the multistep distillation, we reduce the micro step $s$ to 1, resulting in $N$-step sampling without CFG, which still follows the design of StreamDiT for streaming. Moreover, it powers up a significant speed-up that enables real-time video generation. 

%% file: sec/6_exp.tex
\section{Experiments}

\subsection{Implementation Details}

We finetune a T2V model~\cite{moviegen} with 4B parameters using the architecture introduced in~\cref{sec:model}. The latent size is $[16,64,64]$. With a $[4\times,8\times,8\times]$ TAE, the base model generates 64 frames at 512p resolution.

Then we adapt the model for streaming video generation. Our StreamDiT training has three stages: task learning, task generalization, and quality fine-tuning. In the first stage, we use a small amount of high-quality video data (3K videos) with a large learning rate ($1e-4$) to adapt the original T2V into a video streaming model. The second stage involves further training on the pretraining dataset (2.6M videos) with a small learning rate ($1e-5$) to improve generalization for video streaming. In the final stage, we finetune the model on the high-quality video dataset with a small learning rate ($1e-5$) to optimize output quality. Each stage is trained with 10K iterations on 128 NVIDIA H100 GPUs. 

To achieve real-time inference, we additionally apply the multistep distillation described in~\cref{sec:distill}. We found a partitioning scheme $c=2,s=16,N=8$ is good for our multistep distillation considering both quality and efficiency, so we used it for teacher model inference. The distillation is also conducted on the 3K high-quality video dataset and 64 NVIDIA H100 GPUs for 10K iterations. After distillation, the micro step is reduced to 1, and the total number of sampling steps is 8. For more training details, please refer to the Appendix. 

\begin{table}
    \centering
    \resizebox{\columnwidth}{!}{
    \begin{tabular}{lccccccc}
        \toprule
        & \makecell{Subject \\ Consistency} 
        & \makecell{Background \\ Consistency} 
        & \makecell{Temporal \\ Flickering} 
        & \makecell{Motion \\ Smoothness} 
        & \makecell{Dynamic \\ Degree} 
        & \makecell{Aesthetic \\ Quality} 
        & \makecell{Quality \\ Score} \\
        \midrule
        ReuseDiffuse & 0.9501 & 0.9615 & 0.9838 & 0.9912 & 0.2900 & 0.5993 & 0.8019 \\
        FIFO & 0.9412 & 0.9576 & 0.9796 & 0.9889 & 0.3094 & 0.6088 & 0.7981 \\
        Ours & 0.9622 & 0.9625 & 0.9671 & 0.9861 & 0.5240 & 0.6026 & \textbf{0.8185} \\
        Ours-distill & 0.9491 & 0.9555 & 0.9649 & 0.9831 & 0.7040 & 0.5940 & 0.8163 \\
        \bottomrule
    \end{tabular}
    }
    \caption{VBench quality metrics of our evaluation. Our models outperform others, and our distilled model is close to our teacher model.}
    \label{tab:eval}
\end{table}

\subsection{Evaluation}
We compare our method with ReuseDiffuse~\cite{gu2023reusediffuse} and FIFO-Diffusion~\cite{Kim2024FIFO} for streaming generation of long videos. To make a fair comparison, we implemented their methods on our base model, so they share the same visual quality. We select 50 prompts from the evaluation dataset of Movie Gen~\cite{moviegen} that are suitable for long videos. We adopt VBench~\cite{huang2023vbench} quality metrics for quantitative evaluation.

As shown in~\cref{tab:eval}, our models outperform others with high quality scores. All methods have similar aesthetic quality and imaging quality scores. This is because they are derived from the same base model. ReuseDiffuse and FIFO achieve higher temporal consistency and motion smoothness. However, by examining their generated videos, the content is more static. This is also reflected in the dynamic degree column, where our models are much better. 

\begin{figure}
    \centering
    \includegraphics[width=\linewidth]{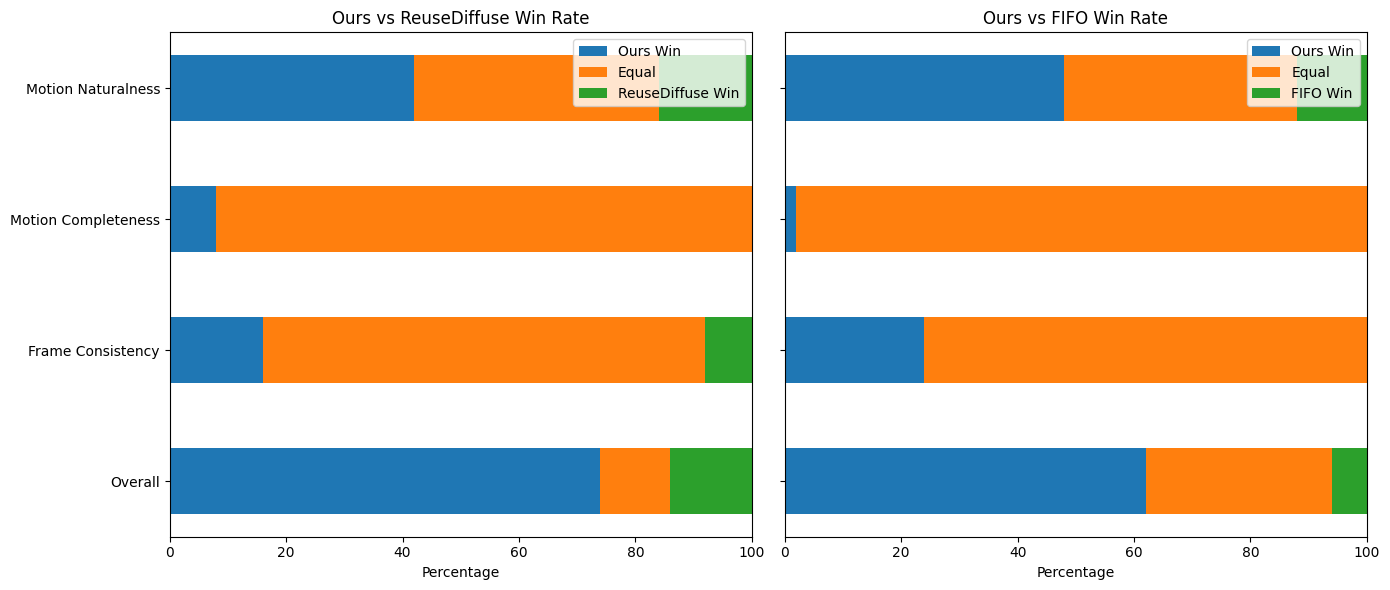}
    \caption{Human evaluations of our method compared with others, where our model shows higher win rates across all axes.}
    \label{fig:user_study}
\end{figure}

In addition to VBench metrics, we also conduct human evaluations following the guidance in~\cite{moviegen}. A human evaluation compares two model results on the same prompts side-by-side. Annotators were asked to evaluate pairs of video samples to determine which one was superior or if they were of comparable quality along several axes. The evaluation criteria encompassed four aspects: overall quality, frame consistency, motion completeness, and motion naturalness. For these evaluations, we employ the same set of 50 prompts used in the VBench evaluation, generating 8-second videos at a resolution of 512p for each prompt. As shown in~\cref{fig:user_study}, our proposed method surpasses existing approaches across all evaluation metrics. 

\begin{figure*}
    \includegraphics[width=\textwidth]{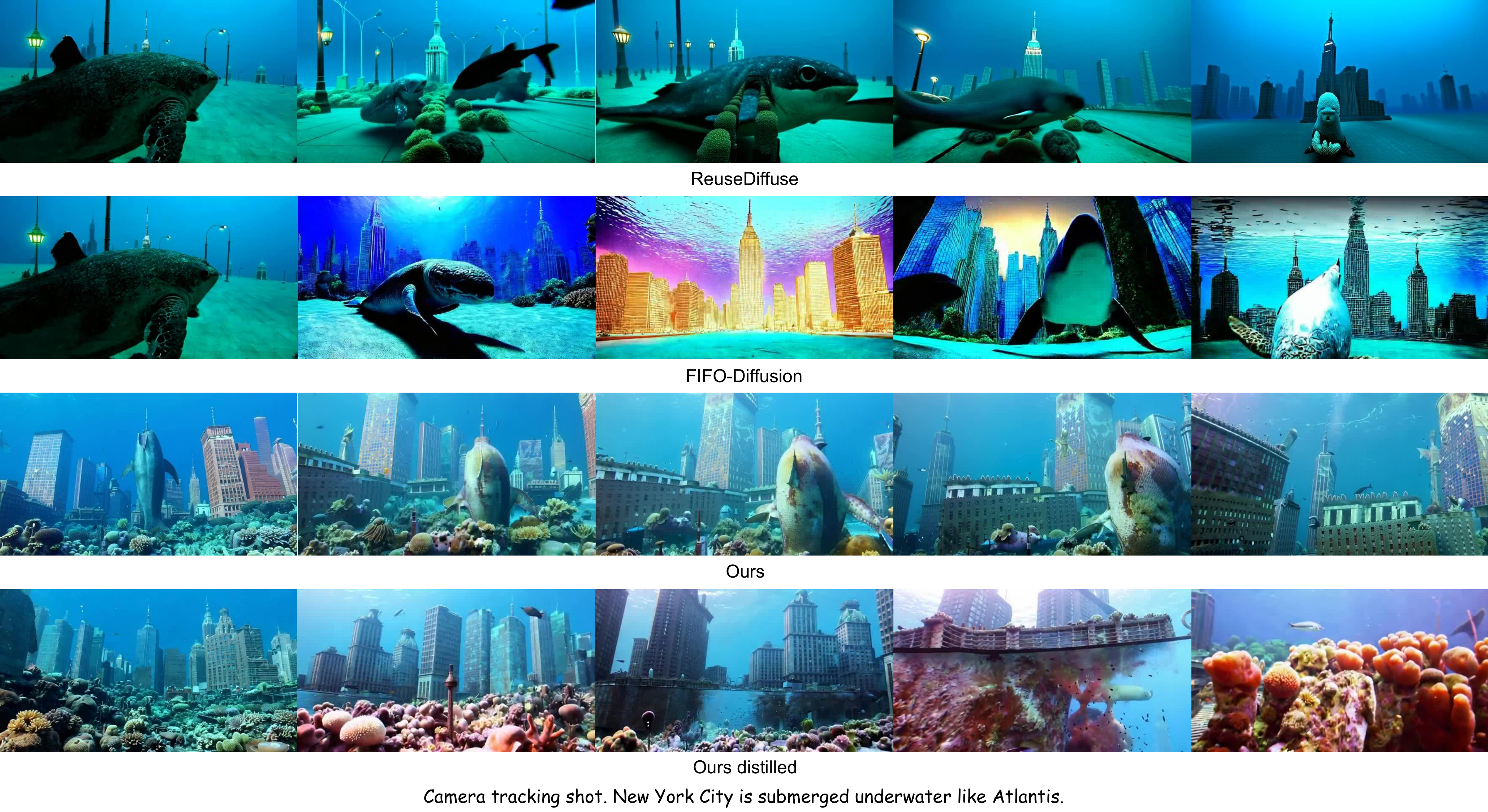}
    \caption{Visual results selected from our evaluation. Our models show better consistency and higher quality than others. Our distilled model has a similar quality with our teacher model.}
    \label{fig:compare}
\end{figure*}

We show some selected frames from generated videos in~\cref{fig:compare}. The videos are one-minute long. We can see that our models have more consistent content with more motions, while others are static. This observation aligns with the VBench metrics.

\begin{table}[]
    \centering    \resizebox{\columnwidth}{!}{
    \begin{tabular}{c|c|c|c|c|c}
    \toprule
       Chunk size & [1] & [1,2] & [1,2,4] & [1,2,4,8] & [1,2,4,8,16] \\ \midrule
       Quality score & 0.8129 & 0.8100 & 0.8080 & 0.8076 & \textbf{0.8144} \\ \bottomrule
    \end{tabular}}
    \caption{Ablation of mixed training. Chunk size 1 represents the Progressive AR Diffusion~\cite{xie2024progressiveautoregressivevideodiffusion}, and chunk size 16 represents the original T2V without streaming. A mixed training with all chunk sizes show the best quality score.}
    \label{tab:ablation}
\end{table}

\subsection{Ablation Study}
We conduct an ablation study to analyze the mixed training discussed in~\cref{sec:buflow}. Specifically, we train models with different mixing schemes as shown in~\cref{tab:ablation}. Chunk size 1 represents the Progressive AR Diffusion~\cite{xie2024progressiveautoregressivevideodiffusion} implemented on our model. Recall that our base model has 16 latent frames, so chunk size 16 is the basic T2V. After the models are trained, we generated videos using chunk size $c=1$, since this is the case covered by all models. Among them, we see that the mixture of all chunk sizes achieves the best quality score, although the inference is biased to chunk size 1 as being set in inference. Please note that we use a different inference scheme in~\cref{tab:eval}, so the numbers of our model are different. The first one with chunk size 1 and no mixing achieves the second-best quality score, indicating an overfitting to this case.

\subsection{Applications}

\begin{figure*}[!ht]
    \includegraphics[width=\textwidth]{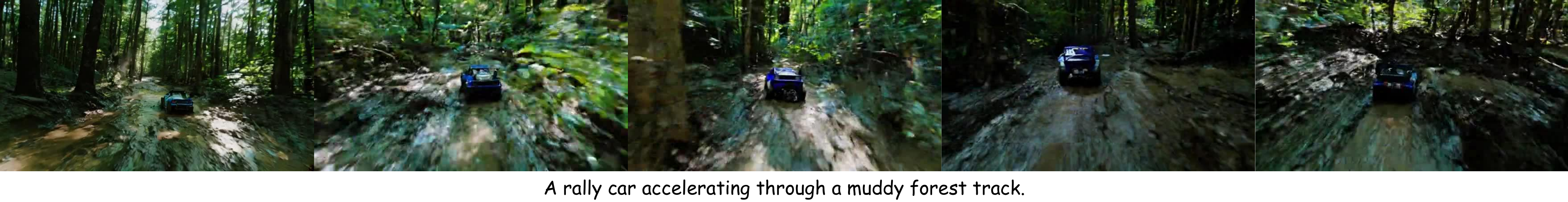}
    \caption{Real-time video streaming. With real-time streaming generation, our model can potentially work as a game engine.}
    \label{fig:game}
\end{figure*}

\textbf{Real-Time Streaming:}
Our distilled model has a chunk size 2 and 8 sampling steps without CFG. We benchmark its performance on one H100 GPU. It takes 482 ms for one denoising step to generate 2 latent frames and 8 video frames after a $4\times$ TAE. The latency of text encodering and TAE decoding are negligible. Thus, our distilled model can reach real-time performance at \textbf{16 FPS}. \cref{fig:game} shows frames from a one-minute long video generated in real-time. With real-time streaming generation, our model can potentially work as a game engine.

\begin{figure*}[!ht]
    \includegraphics[width=\textwidth]{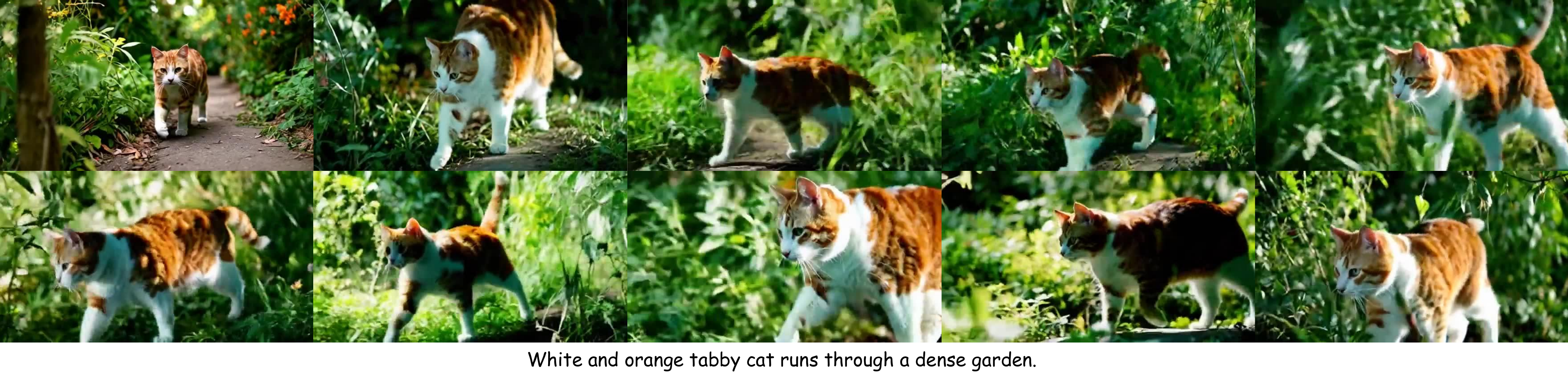}
    \caption{Infinite streaming. We generate a 5-minute video to demonstrate the potential for infinite streaming.}
    \label{fig:long}
\end{figure*}

\textbf{Infinite Streaming:}
To explore the ability of infinite streaming, we try our distilled model to generate a video over 5 minutes long. As shown in~\cref{fig:long}, the video content and quality are consistent after a very long generation, demonstrating the potential for infinite streaming. 

\begin{figure*}[!ht]
    \includegraphics[width=\textwidth]{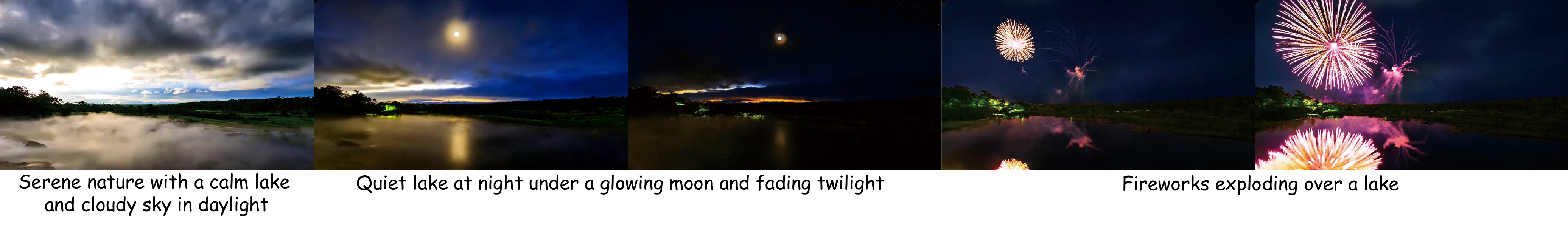}
    \caption{Interactive video streaming. The user can enter prompts to navigate the video generation on the fly.}
    \label{fig:story}
\end{figure*}

\textbf{Interactive Streaming:}
Additionally, we showcase the interactive storytelling capabilities of our model using a sequence of semantically related prompts. The qualitative results in~\cref{fig:story} demonstrate that our model effectively controls video events interactively, based on user-specified prompts.

\begin{figure}[!ht]
    \includegraphics[width=\linewidth]{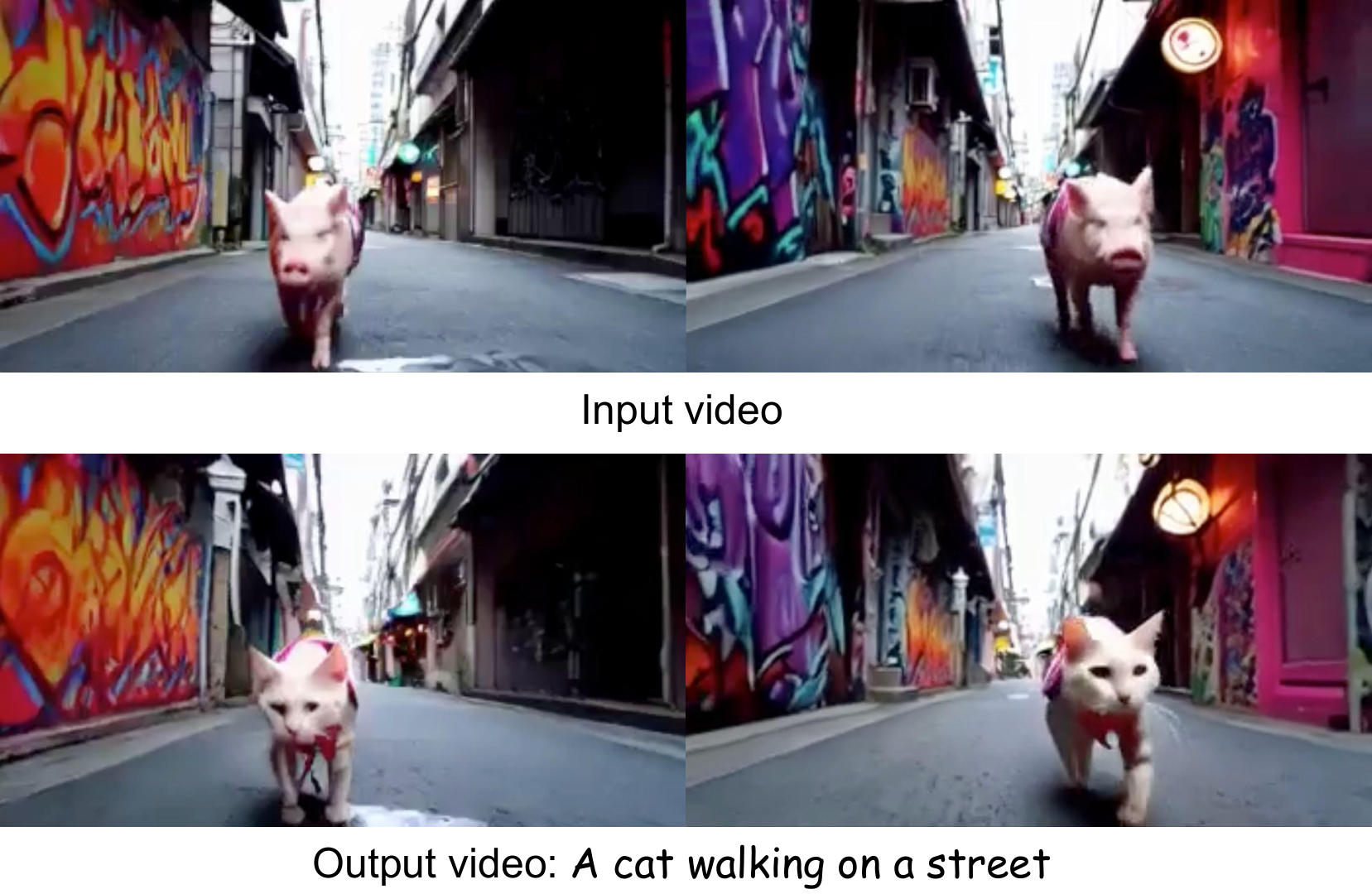}
    \caption{Video-to-video streaming. The top row is input and the bottom row is output, where a pig is modified to a cat by prompt.}
    \label{fig:v2v}
    \vspace{-10pt}
\end{figure}

\textbf{Video-to-Video Streaming:}
Our model can also be applied to real-time video-to-video tasks.  The video-to-video process is based on the noise addition and denoising strategy proposed in SDEdit~\cite{mengSDEditGuidedImage2022}. First, we add noise to the original input video and then denoise the noisy video with new prompt guidance. We achieve significant content editing while maintaining good temporal consistency.
The visual results shown in~\cref{fig:v2v} highlight the model’s powerful video editing capabilities. In this example, a pig in the input video is modified to a cat in the output video, while the background is preserved.

%% file: sec/7_conclu.tex
\section{Conclusion and Limitations}

In this work, we present StreamDiT for streaming video generation. It includes a novel training framework and efficient modeling that can run in real-time. StreamDiT training is based on flow matching with a moving buffer. We design generalized partitioning that unifies different schemes. Through experiments, we show that mixed training with different schemes can improve the consistency and quality of generations. To achieve real-time performance, we train an efficient StreamDiT model based on time-varying DiT with window attention. We further distill the model to 8 sampling steps, using the proposed multistep distillation tailored for StreamDiT. Our model achieves high temporal consistency across frames, addressing critical challenges in long-form and interactive video applications. 

At the end, we discuss two limitations of our StreamDiT-4B, regarding model capacity and context length. StreamDiT-4B has a moderate model size, which limits the quality of basic T2V generation. As a result, we noticed artifacts in some of the generated videos.
To verify the capacity limitation and the scalability of our method, we also applied StreamDiT to a 30B-parameter model with a light-weight training process. We found that StreamDiT-30B shows much better quality in terms of visual aesthetic, content, and low artifacts. Please refer to the Appendix and project website for the results. Another limitation of StreamDiT-4B is the short context length. Flaws may be observed whereby out-of-context objects exhibit altered visual appearance after they re-enter the video sequence. We anticipate that this limitation can be mitigated by employing an extended context window in conjunction with a key–value (KV) cache during inference, thereby enabling more efficient computational acceleration. Therefore, the two limitations are not fundamental to the StreamDiT method, and can be improved with future work.

%% file: sec/8_acknowledge.tex
\section{Acknowledgment}
We would like to express our sincere gratitude to Yen-Cheng Liu, Luxin Zhang, and Tao Xu for their valuable discussions and insightful contributions, which greatly enriched our work.

%% file: sec/appendix.tex
\clearpage
\setcounter{page}{1}
\maketitlesupplementary

\section{Inference Details}
\label{sec:inferencee_details}
\begin{figure}[!t]
    \includegraphics[width=\linewidth]{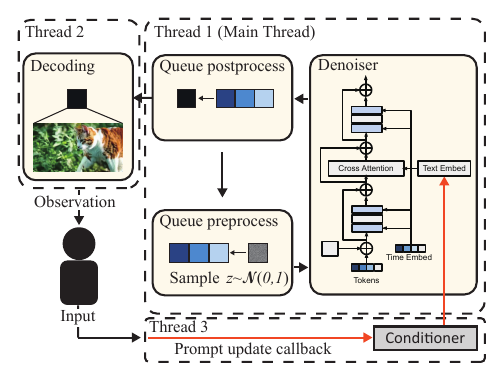}
    \caption{Interactive inference pipeline of StreamDiT: To decrease latency, generative models, decoder and text encoder are in separate process.}
    \vspace{-5pt}
    \label{fig:inference_pipeline}
\end{figure}

StreamDiT is specifically designed to achieve real-time responsiveness and interactivity, and its inference pipeline is structured accordingly. An overview of this pipeline is provided in~\cref{fig:inference_pipeline}. In the main thread~(Thread 1), the system performs the denoising operation, refills the stream queue, and emits denoised video frames from the queue to forward them to a separate decoder thread~(Thread 2). This decoder thread runs concurrently, decoding the latent video frames to actual video frames. These resulting frames are then rendered in real time, allowing users to observe the changes immediately.

\begin{figure}[!t]
    \includegraphics[width=\linewidth]{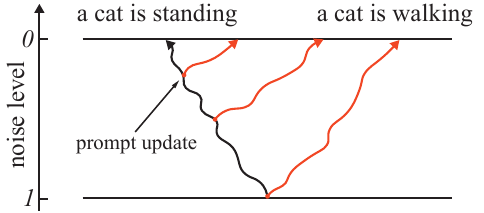}
    \caption{Denoising trajectory change with text guidance update:
As denoising progresses toward the final stages, it becomes increasingly difficult to deviate from the outcome dictated by the original text guidance.}
   \vspace{-15pt}
    \label{fig:prompt_update}
\end{figure}

Additionally, a prompt callback function operates continuously on another thread~(Thread 3), listening for new user prompts in real time. When a user provides a new prompt, it is converted into text embedding by text encoders, and the embedding is sent to the DiT thread to update the existing embedding. Subsequent denoising steps then use this updated embedding through a cross-attention mechanism, changing the direction of text guidance dynamically~(\cref{fig:prompt_update}). This design enables users to interactively influence and modify video content in real time through prompt inputs. 

In StreamDiT, since information from both preceding and succeeding video segments is always present in the context, it is essential to consider not only the explicit text guidance but also the implicit influence of video guidance. When the chunk size is small, the noise level difference between adjacent blocks in the Stream Queue decreases, amplifying the effect of video guidance. Therefore, to allow for larger content transformations, such as morphing, a larger chunk size is preferable. Conversely, if the goal is to maintain semantically continuous changes, such as variations in the walking direction of a character, a smaller chunk size is more suitable.

\begin{figure*}[!ht]
    \includegraphics[width=\textwidth]{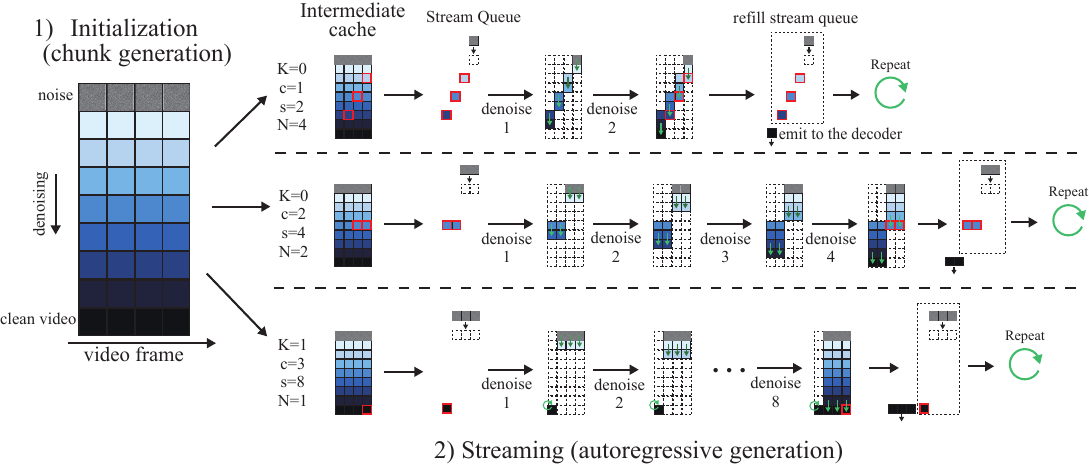}
    \caption{In the T2V scenario, StreamDiT first performs standard chunk generation to prepare intermediate latent cache. Once the intermediate cache is fully populated, the appropriate blocks are retrieved to construct the initial Stream Queue. Different inference configurations can be activated, provided that those configurations were included during mixed chunk training.}
    \label{fig:generation_variant}
\end{figure*}

To perform stream denoising, the Stream Queue is filled with latent video frames that vary gradually in noise level. In the case where an initial video is provided in advance, such as in video-to-video scenarios, the video can first be encoded and then filled into the Stream Queue by adding Gaussian noise with appropriate stepwise noise levels. However, in the case of text-to-video generation, as shown in~\cref{fig:generation_variant}, the process starts with chunk generation using a standard T2V model. During this stage, the intermediate video latents at each denoising step must be cached. Afterward, from the cached intermediate latents, those corresponding to the appropriate noise levels and video frames are selected in accordance with the StreamDiT inference configuration, and used to populate the Stream Queue. Once the queue is prepared, infinite-length video generation becomes possible by autoregressively and continuously performing stream denoising. Furthermore, as illustrated in the bottom row of~\cref{fig:generation_variant}, when the number of stream chunks $N$ is set to one and one or more reference frames are used~($K\geq1$), conventional chunk-based video extension is also possible. Because of StreamDiT's highly flexible unified architecture, it is possible to choose the chunk size and block size mixed during training according to the intended use. This enables a single model to flexibly switch between various inference patterns during deployment.

\section{Design Choices and Training Details}
\label{sec:training_details}
Our design process begins by defining the global context length \(B\) of the base text-to-video (T2V) model. Once \(B\) is established, we determine a corresponding micro-step budget, enabling the informed selection of an appropriate chunk size \(c\) to meet latency constraints. The choice of chunk size critically balances responsiveness and visual fidelity; larger chunk sizes yield more stable and higher-quality outputs but incur increased latency.
To ensure flexibility across various inference scenarios, we adopt mixed-chunk training, which minimally impacts overall performance. Notably, incorporating the full-context chunk size (\(c = B\)) into mixed-chunk training substantially improves output quality by preserving key characteristics of the original T2V model, as demonstrated in~\cref{tab:ablation}.
For our distilled real-time model, we select a chunk size of \(c = 2\) to achieve optimal performance at 16~FPS. Choosing a smaller chunk size (\(c = 1\)) would reduce performance to approximately 8~FPS in a step-distilled scenario. The micro-step size \(s\) is not tuned independently; rather, it is directly determined by the chosen chunk size and the total number of denoising steps (\(T = s \times N\)).

In this paper, we trained StreamDiT using a partitioning strategy based on a \textit{linear noise schedule}. However, depending on the denoising scenario, it may be beneficial to adopt a \textit{nonlinear noise schedule} during inference. For instance, stronger denoising may be preferred in earlier frames, while detailed reconstruction might be prioritized in later ones.
To enhance the performance of stream denoising under such conditions, it is crucial that the noise distribution during training closely matches the intended inference-time noise schedule~(\cref{fig:training_partitioned}). To achieve this, we define a general noise scheduling function $\gamma(t)$, mapping the normalized time domain $[0,1]$ to the noise level range $[0,1]$. Here, $t$ represents normalized temporal positions, such as frame indices or timestamps. We partition the time domain into $N$ equal segments, and within each segment $\left( \frac{i-1}{N}, \frac{i}{N} \right]$, we randomly select a time point $t_i$ from a uniform distribution. The noise level at this sampled point, $\gamma(t_i)$, is then applied uniformly across the entire interval. The resulting stepwise noise function $\hat{\gamma}(t)$ is expressed as:

\begin{figure}[!t]
    \includegraphics[width=\linewidth]{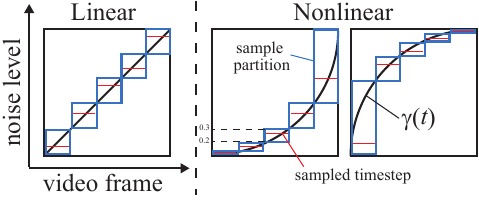}
    \caption{Partitioned Noise Training: The partitioning strategy can be defined by an arbitrary function—linear or nonlinear—that aligns with the chosen noise scheduling strategy.}
    \label{fig:training_partitioned}
       \vspace{-10pt}
\end{figure}

\begin{figure*}[!ht]
    \includegraphics[width=\linewidth]{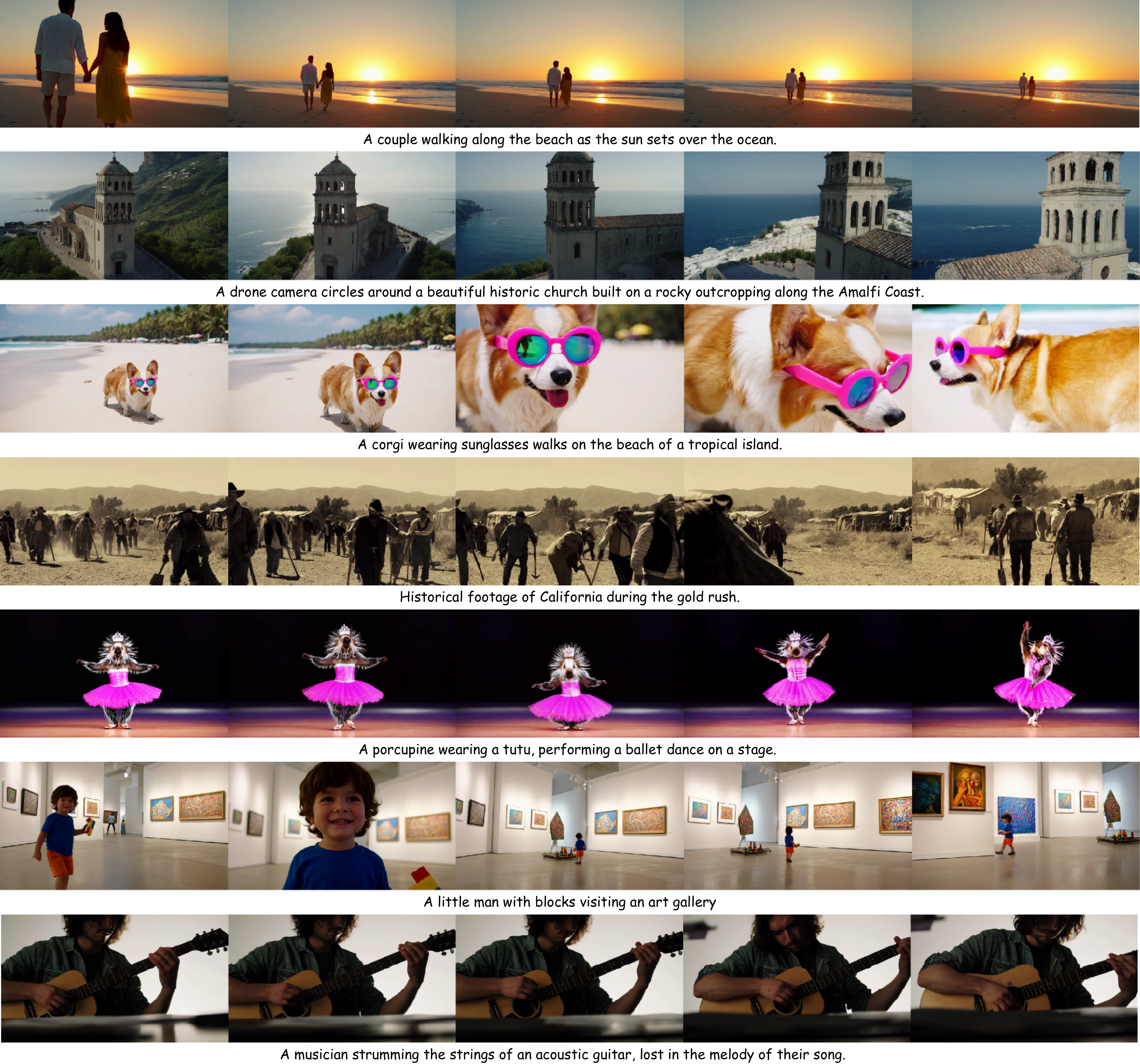}
    \caption{High-quality long videos generated by StremDiT-30B, demonstrating the scalability of our method.}
    \vspace{-10pt}
    \label{fig:scalability}
    \vspace{-5pt}
\end{figure*}

\begin{align}
\hat{\gamma}(t) &= \sum_{i=1}^{N} \gamma(t_i) \cdot \mathbf{1} \left[ 
t \in \left(\frac{i-1}{N}, \frac{i}{N} \right] \right], \nonumber \\
&\text{where } t_i \sim \text{Uniform} \left( \left[\frac{i-1}{N}, \frac{i}{N}\right] \right).
\end{align}
This approach enables us to approximate various nonlinear noise schedules with an easily implementable step function. For example, a linear schedule corresponds to $\gamma(t) = t$, while exponential schedules can be modeled as $\gamma(t) = t^k$ with $k > 1$. Thus, our formulation provides flexibility to match different inference-time behaviors.

Furthermore, StreamDiT involves latent representations with different noise levels interacting within a shared context. Due to this design, the model is sensitive to the chosen noise scheduling strategy and the context length. Therefore, careful consideration is required when setting these parameters. For instance, if most frames are assigned high noise levels, these frames will convey significantly less information compared to frames with lower noise. This scenario effectively reduces the usable context size, restricts information flow across frames, and can increase the difficulty of training. Consequently, improper design of noise schedules can negatively impact the denoising quality and temporal consistency of the model’s outputs. Hence, selecting an appropriate noise scheduling strategy is crucial for achieving optimal model performance.

\begin{figure*}[!ht]
    \includegraphics[width=\linewidth]{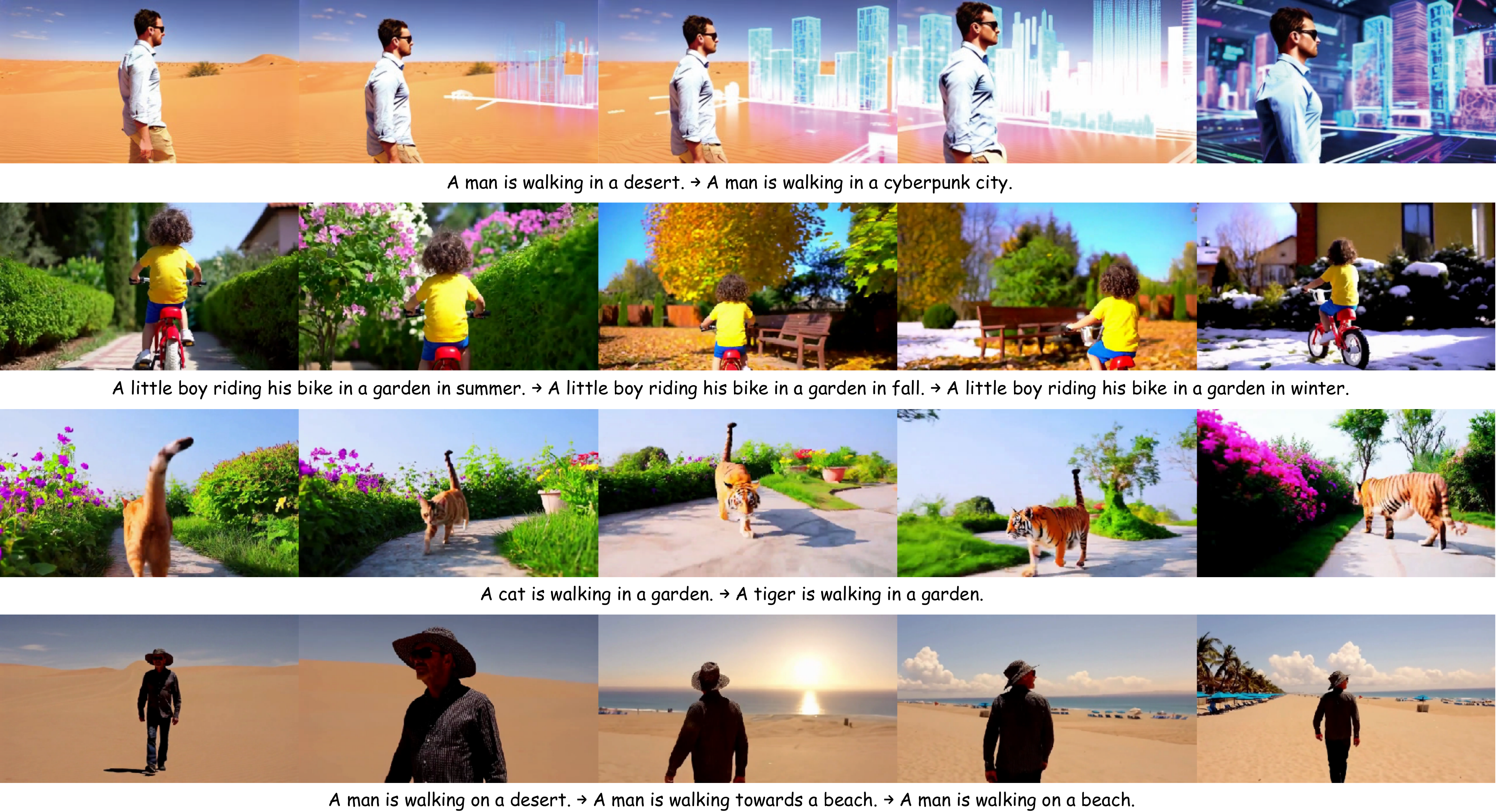}
    \caption{Sequential storytelling prompts can mitigate repetitive content and enable dynamic contents change.}
    \label{fig:prompt_sequence}
\end{figure*}

\begin{table*}[t]
    \centering
    \begin{tabular}{c|c|c||c|c|c|c|c|c|c}
    \toprule
        \multicolumn{2}{c|}{Config} & Quality & Subject & Background & Flickering & Motion & Dynamic & Aesthetic & Imaging \\
        \midrule
        \multirow{8}{4em}{Reference frames} & fixed 1 & 0.8175 & 0.9668	&0.9769	&0.9813&	0.9910&	0.4767&	0.5500&	0.6683\\
         & fixed 2 & 0.8202 &0.9643	&0.9763	&0.9810	&0.9911	&0.5533	&0.5482	&0.6536\\
         & fixed 4 & 0.8151 & 0.9634	&0.9748	&0.9736	&0.9886	&0.5523	&0.5292	&0.6713\\
         & fixed 8 & 0.8222 & 0.9664	&0.9768	&0.9809	&0.9913	&0.5320	&0.5538	&0.6680\\
         & mixed 1 & 0.8228 & 0.9652& 	0.9765	& 0.9797& 	0.9907& 	0.5733& 	0.5421& 	0.6704\\
         & mixed 2 & 0.8244 & 0.9657	&0.9766	&0.9787	&0.9903	&0.6067	&0.5410	&0.6685\\
         & mixed 4 & 0.8230 & 0.9647	&0.9769	&0.9787	&0.9902	&0.5880	&0.5413	&0.6691\\
         & mixed 8 & 0.8246 &0.9654	&0.9761	&0.9784	&0.9901	&0.6120	&0.5418	&0.6688\\
        \midrule
        \multirow{3}{4em}{T2V mix ratio} & 0.1 & 0.8187 & 0.9690	&0.9796	&0.9823	&0.9909	&0.4852	&0.5437	&0.6693\\
        & 0.3 & 0.8218 & 0.9611	&0.9761&	0.9817	&0.9913&	0.5600	&0.5497	&0.6605\\
        & 0.5 & 0.8186 & 0.9626&	0.9752	&0.9770&	0.9893&	0.5778	&0.5396	&0.6600\\
        \midrule
        \bottomrule
    \end{tabular}
    \caption{Full VBench scores of our ablation studies on the 30B model.}
    \label{tab:ablation_all}
\end{table*}

\section{More Results}
\label{sec:more_results}
To evaluate the capabilities in generating higher-quality videos, we further fine-tuned the 30B model from Movie Gen~\cite{moviegen}. While the 30B model is not available for real-time generation at the present, it effectively supports the creation of exceptionally high-quality videos, as illustrated in~\cref{fig:scalability}. Our model consistently produces extended videos characterized by impressive visual quality, content coherence, and diverse scenes aligned accurately with provided text prompts. 

Additionally, ablation studies on the 30B model using VBench are summarized in~\cref{tab:ablation_all}. The observed quality scores exhibit minimal variation, underscoring the stability and robustness of our method across various configurations.

Due to the inherent context-length limitations of base T2V models, even when augmented by autoregressive denoising improvements such as StreamDiT, the effective temporal context remains restricted. Consequently, repeatedly using the same prompt results in increasingly repetitive content, limiting the diversity of the generated videos. To address this limitation, we propose using a sequence of different story telling prompts during inference. This strategy maintains temporal coherence while allowing dynamic variation in the generated content~(\cref{sec:inferencee_details}). As shown in~\cref{fig:prompt_sequence}, sequential storytelling prompts significantly mitigate repetitive visual patterns and enhancing the feasibility in long-form video generation tasks. This approach facilitates the production of coherent videos with creative content, smooth object transformations, and seamless scene transitions.

\begin{figure}[!t]
    \includegraphics[width=\linewidth]{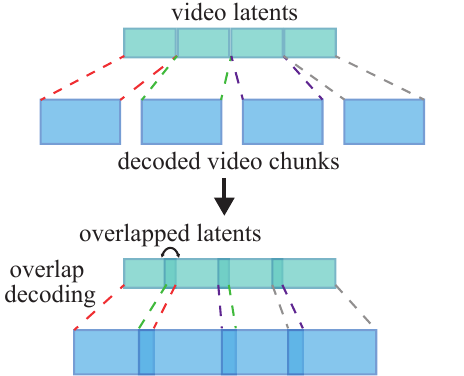}
    \caption{Overlap decoding}
    \label{fig:overlap_decoding}
    \vspace{-10pt}
\end{figure}

\section{Limitations}
\label{sec:Limitation}
As described in~\cref{sec:inferencee_details}, the effective context length of StreamDiT fundamentally depends on the base T2V model, and thus lacks long-term memory. When content falls outside the short-term memory window (\ie context length) of StreamDiT, the associated information is likely to be lost. This can lead to issues such as identity mismatches in a person’s face, or background inconsistencies when the camera makes a full rotation. However, since StreamDiT is orthogonal to additional long-term memory mechanisms, it is possible to address this issue in the future by combining it with long-term memory architectures such as State Space Models like Mamba~\cite{gu2023mamba}.

Furthermore, with the current decoding strategy of StreamDiT, although the video frames are smoothly connected at the latent level, because video latent chunks are sequentially emitted from the Stream Queue and decoded separately, slight seams or flickering may be observed between decoded chunks in the final video. This occurs because a smooth connection at the latent level does not guarantee a seamless reconstruction in the decoded video. As a potential solution shown in~\cref{fig:overlap_decoding}, when decoding, one could concatenate the end of a cached previous video latent to the beginning of a newly emitted video latent from the Stream Queue to create an extended chunk, which is then decoded. This overlapping strategy helps to reduce the appearance of seams.

The improvement of StreamDiT with those additional approaches is left as future work.